\documentclass[5p,twocolumn,number]{elsarticle}

\usepackage[T1]{fontenc}
\usepackage{newtxtext,newtxmath}

\usepackage{multirow}
\usepackage{float}
\usepackage{algorithm}
\usepackage{algpseudocode}
\usepackage{graphicx}
\usepackage{textcomp}
\usepackage{xcolor}
\usepackage{hyperref}
\usepackage{tabularray}
\usepackage{adjustbox}
\usepackage{booktabs}
\usepackage{flushend}
\usepackage{stfloats}

\hypersetup{
 colorlinks=true,
 linkcolor=black,
 filecolor=black,
 urlcolor=blue,
 citecolor=black,
}

\biboptions{sort&compress}

\journal{Computer Networks}
\makeatletter
\def\ps@pprintTitle{%
  \let\@oddhead\@empty
  \let\@evenhead\@empty
  \let\@oddfoot\@empty
  \let\@evenfoot\@empty
}
\makeatother

\newcommand{\authorentry}[3]{%
\vspace{2pt}\par\noindent
\begin{minipage}[t]{0.78in}%
\vspace{0pt}%
\includegraphics[width=0.78in,height=0.96in,clip,keepaspectratio]{#1}%
\end{minipage}%
\hspace{4pt}%
\begin{minipage}[t]{\dimexpr\columnwidth-0.78in-6pt\relax}%
\vspace{0pt}%
\textbf{#2}\space #3%
\end{minipage}\par\vspace{2pt}%
}

\begin{document}

\begin{frontmatter}

\title{SCALE: Scalable Cross-Attention Learning with Extrapolation for Agentic Workflow Scheduling}

\author[bnu]{Zhifei Xu}
\author[bnu]{Jierui Lan}
\author[bnu]{Zixuan Liang}
\author[bnu]{Aiji Liang\corref{cor1}}
\ead{liangaiji@mail.bnu.edu.cn}
\author[bnu]{Jinxi He}

\cortext[cor1]{Corresponding author}

\address[bnu]{Faculty of Arts and Sciences, Beijing Normal University, Zhuhai 519087, China}

\begin{abstract}
Agentic Large Language Model (LLM) systems decompose complex tasks into workflow Directed Acyclic Graphs (DAGs) whose primitives must be scheduled on heterogeneous clusters. Existing deep reinforcement learning (DRL) schedulers are tied to a fixed cluster size and require retraining whenever the number of servers changes. We propose \textbf{SCALE} (\textbf{S}calable \textbf{C}ross-\textbf{A}ttention \textbf{L}earning with \textbf{E}xtrapolation), a DRL scheduler that generalizes to unseen cluster scales without fine-tuning. SCALE employs a cross-attention pointer network where task features query against server features, so the architecture accepts any number of servers by construction. We observe, however, that permutation-invariant architecture alone does not guarantee good performance at new scales—the attention feature undergoes distribution shift as the server count grows. To counter this, we introduce \textbf{S}tructured \textbf{R}epresentation \textbf{R}egularization (SRR): a decorrelation loss combined with a KL penalty toward the standard normal, which keeps feature statistics stable regardless of input size. Trained on 16 nodes and tested directly on 32 and 48 nodes, SCALE reduces average response time by 8.9\% at $N{=}48$ relative to the same architecture without SRR, confirming that explicit regularization is necessary to close the scale-generalization gap.
\end{abstract}

\begin{keyword}
Workflow Scheduling \sep Cross-attention \sep Scalable Reinforcement Learning
\end{keyword}

\end{frontmatter}

\section{Introduction} \label{Introduction}
Agentic Large Language Model (LLM) systems do more than answer questions. They perceive environments, make plans, invoke external tools, and self-correct when execution goes wrong \cite{plaat2025agentic}. A planner inside these systems breaks high-level goals into Directed Acyclic Graphs (DAGs) of primitive operations \cite{kim2023llm}—each node being a tool call, each edge a data dependency. How these primitives get mapped onto compute hardware directly determines response latency and cluster utilization. Unlike single-model inference serving, the cluster here is heterogeneous, its available capacity shifts at runtime, and individual primitives differ greatly in memory and compute needs.

Cloud-based task scheduling has been studied extensively \cite{armbrust2010view}, yet the bulk of that literature addresses MapReduce jobs \cite{dean2008mapreduce} or microservice chains \cite{zhang2021sinan}, workloads whose structure differs substantially from agentic DAGs. Several recent analyses \cite{chaudhry2025murakkab, shen2025batch} observe that conventional schedulers cannot capture DAG-level dependencies together with per-primitive heterogeneity in compute and memory. Agentic tasks compound the difficulty: execution paths are decided on-the-fly and arrivals are stochastic, rendering static resource reservation largely useless \cite{cheng2024slice}. A further obstacle is that current DRL-based schedulers embed the cluster size into their network architecture, so any change in the number of servers forces a full retraining cycle \cite{ma2024efficient}. Designing a scheduler for this setting raises two concrete difficulties.

\textit{Representing a heterogeneous, variable-size state.} The cluster state mixes task-level attributes (compute demand, memory footprint) with server-level attributes (capacity, current load), and the server count $N$ may differ between training and deployment. Standard MLP policy networks \cite{schulman2017proximal} accept fixed-dimension inputs, so they break outright when $N$ changes. Graph Neural Networks (GNNs) can encode task dependencies \cite{liu2024ga}, but merging task and server features into one graph adds substantial computational overhead.

\begin{figure}[t]
    \centering
    \includegraphics[width=\linewidth]{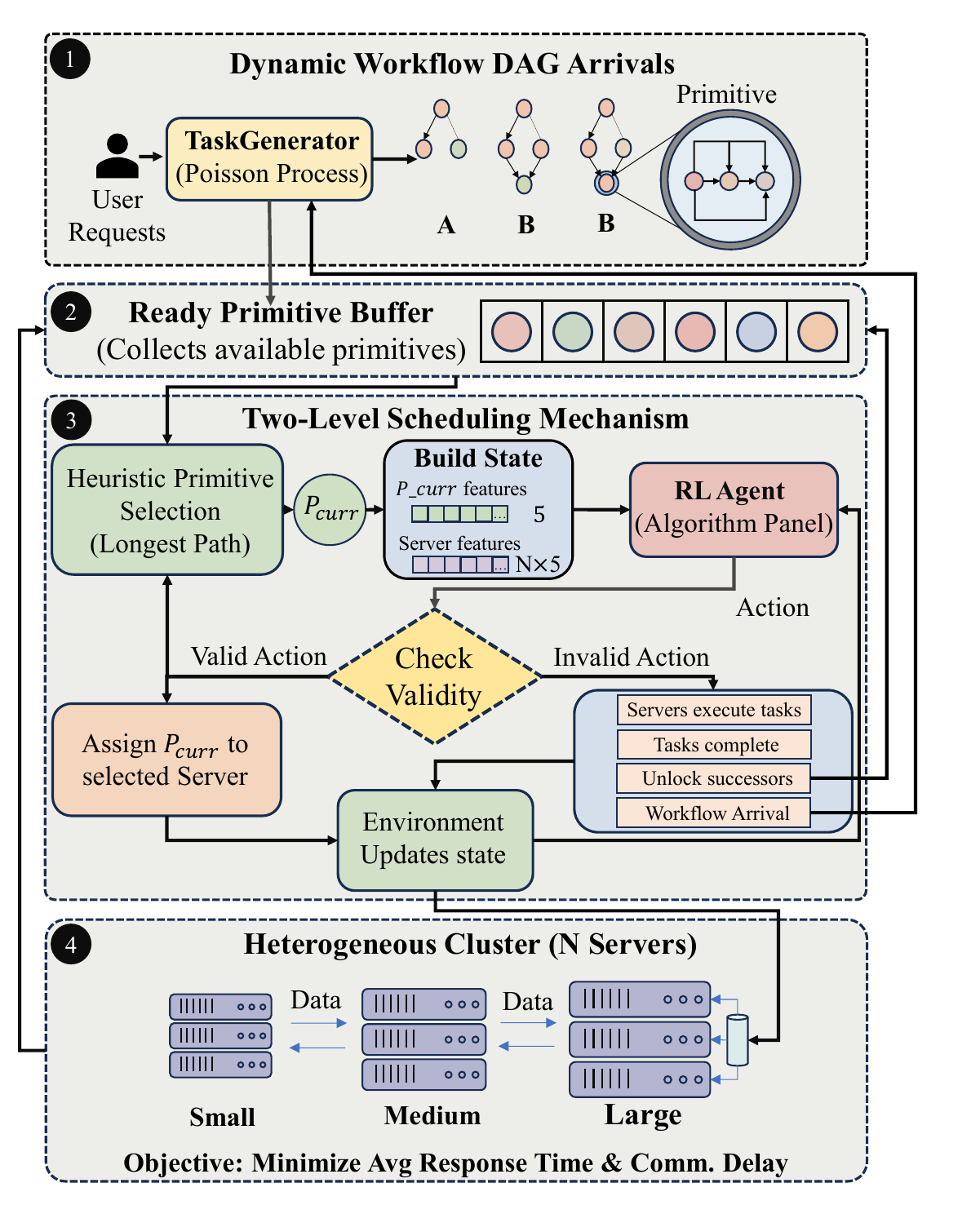}
    \caption{System architecture for agentic workflow scheduling. Workflow DAGs arrive dynamically and are collected in a Ready Primitive Buffer. A two-level scheduler—heuristic longest-path selection followed by an RL agent—assigns primitives to a heterogeneous server cluster, minimizing average response time and communication delay.}
    \label{fig:system_architecture}
\end{figure}

\begin{figure}[t]
    \centering
    \includegraphics[width=\linewidth]{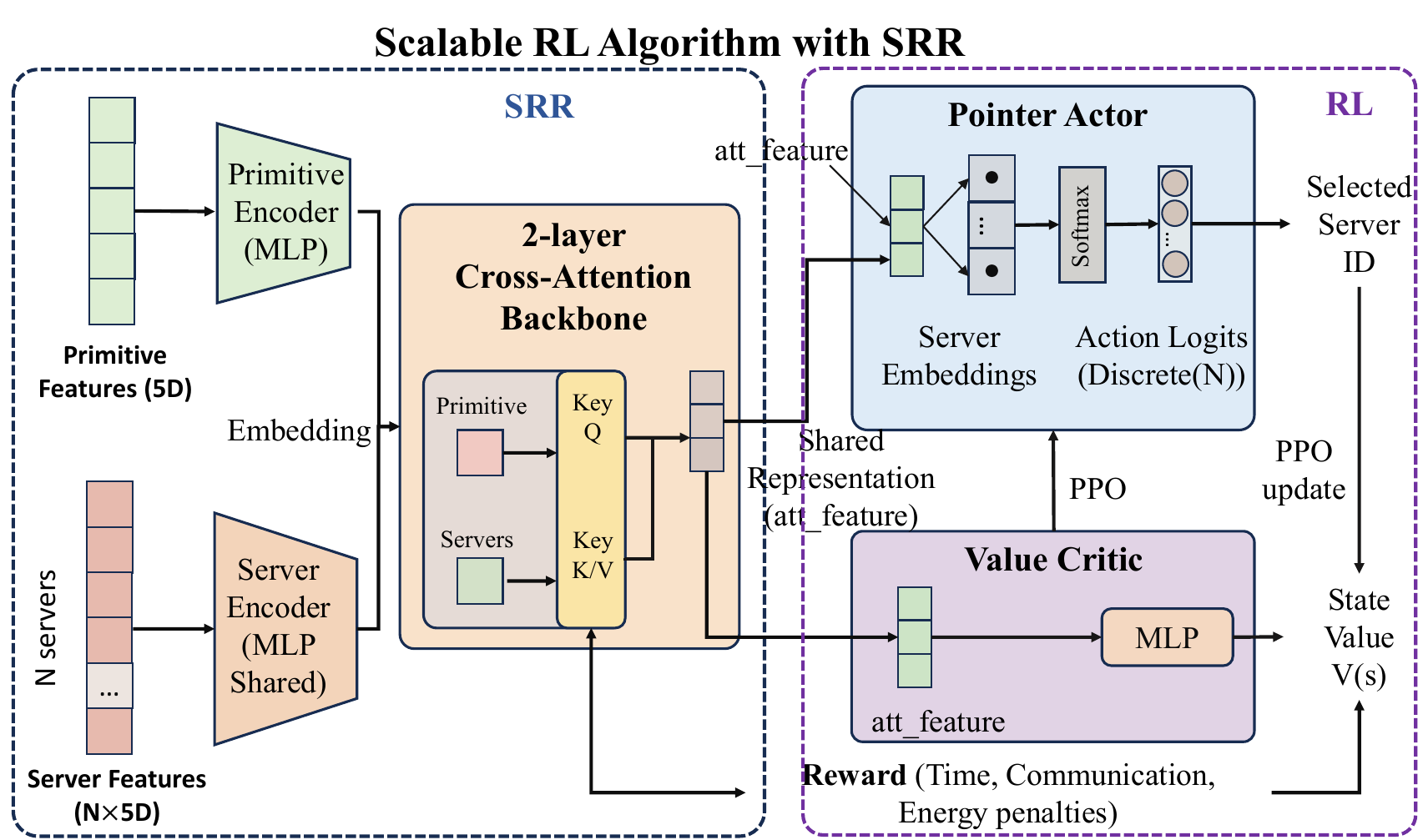}
    \caption{Architecture of the SCALE algorithm with SRR. A 2-layer cross-attention backbone integrates primitive features (queries) with server features (key-value pairs). The resulting representation feeds a Pointer Actor (softmax over server scores) and a Value Critic, both trained via PPO.}
    \label{fig:scalable_rl_srr}
\end{figure}

\textit{Generalizing across cluster scales.} We want to train on a small cluster ($N$=16) and deploy on larger ones ($N$=32, 48, or beyond) without retraining. In practice, a policy learned at one scale degrades at another because the latent features undergo distribution shift. In attention-based architectures \cite{vaswani2017attention}, the variance of the aggregated representation grows with the key-set size, which causes feature dimensions to become redundant and correlated. The policy then becomes brittle once the input set exceeds what was seen during training.

We tackle both issues with \underline{S}calable \underline{C}ross-\underline{A}ttention \underline{L}earning with \underline{E}xtrapolation (SCALE). Figures \ref{fig:system_architecture} and \ref{fig:scalable_rl_srr} show the system overview and algorithm architecture, respectively. The core mechanism is cross-attention: the current task's features serve as the query while all server states form the key-value set, allowing the model to accept any number of servers without modification. On top of this, a pointer network \cite{vinyals2015pointer} computes selection scores through query-key dot products, yielding an output whose dimension equals the current $N$. The main technical contribution is structured representation regularization (SRR)—a decorrelation penalty on the off-diagonal covariance of the attention feature, paired with a KL term that pulls each dimension's marginal toward $\mathcal{N}(0,1)$. SRR prevents the feature statistics from drifting as $N$ grows, so a model trained at $N$=16 can run at $N$=32 or 48 without any fine-tuning.

We evaluate SCALE on a heterogeneous simulated cluster, training at $N$=16 and testing at $N$=32 and 48 with no fine-tuning. We compare with several DRL baselines and ablate SRR's contribution to generalization.

Our contributions are:
\begin{enumerate}
    \item A formalization of agentic workflow scheduling as an MDP over dynamically arriving DAGs on a heterogeneous cluster, jointly capturing task dependencies, resource heterogeneity, and communication costs.
    \item The SCALE algorithm—a cross-attention pointer network augmented with SRR—that achieves zero-shot generalization across cluster sizes.
    \item Experimental evidence that SCALE, trained at $N$=16, retains competitive response time at $N$=32 and 48, whereas the same architecture without SRR degrades noticeably.
\end{enumerate}

The rest of this paper proceeds as follows. Section \ref{Relatedwork} surveys related work. Section \ref{Problem Formulation} formalizes the system model. Section \ref{Algorithm} describes SCALE. Section \ref{Experiment And Result} presents experiments. Section \ref{Discussion} discusses limitations and Section \ref{Conclusion} concludes.

\section{Related Work} \label{Relatedwork}

\textit{Agentic LLM Systems.} An agentic LLM system iteratively decomposes goals, invokes tools, and revises its plan based on intermediate results—going well beyond single-turn generation. AutoGPT \cite{significantgravitas2023autogpt} demonstrated recursive task decomposition driven entirely by an LLM. LangChain \cite{chase2022langchain} introduced reusable abstractions for tool calling, memory management, and chain-of-thought orchestration. MetaGPT \cite{hong2024metagpt} assigned distinct roles to collaborating agents, emulating structured development processes.

From a scheduling standpoint, the critical observation is that these workflows are DAGs. LangGraph \cite{langchain2024langgraph} and AutoGen \cite{wu2023autogen} represent agent interactions explicitly as graphs with conditional branches and parallel paths. Nodes are tool calls or reasoning steps; edges are data dependencies. Different tool calls—code execution, API queries, database lookups—exhibit vastly different latencies and resource footprints, and the number of concurrently active agents fluctuates rapidly. Existing orchestration frameworks, however, assume a fixed compute pool and do not adjust scheduling to a changing cluster. Our method instead treats the server set as variable-size input and learns a policy that adapts to arbitrary cluster configurations.

\textit{Workflow Scheduling.} DAG scheduling on distributed clusters has been studied for decades. The classical list heuristics HEFT and CPOP \cite{1999a} run in $O(v^2 p)$ for $v$ tasks and $p$ processors and yield good makespans, but they require the processor set to be fixed and known ahead of time. Neither can react to nodes joining or leaving at runtime.

DRL introduced learnable scheduling policies. DeepRM \cite{mao2016deeprm} used a fully connected policy network to bin-pack jobs onto machines. Decima \cite{mao2019decima} went further by encoding DAG structure with a GNN, outperforming hand-tuned heuristics on dataflow workloads. Both, however, are trained for a specific cluster size—their policy networks have a fixed output dimension and cannot be reused when $N$ changes. Multi-objective variants share this limitation. Our work targets elastic clusters where nodes may be added or removed at runtime due to auto-scaling, preemption, or failure, demanding a scheduler that generalizes across sizes without retraining.

\textit{Scalable RL for Variable-size Problems.} The fundamental difficulty in applying RL to scheduling is that both the task graph and the server set vary in size across instances. Fixed-dimensional MLP policies fail whenever the problem size changes. Two lines of work address this.

One line relies on permutation-invariant encoders. GCNs \cite{kipf2017gcn} handle graph-structured inputs and have been incorporated into scheduling policies. For unordered sets such as available machines, Deep Sets \cite{zaheer2017deep} and Set Transformers \cite{lee2019set} offer architectures whose output is independent of input ordering.

The other line exploits attention for combinatorial optimization. The Attention Model \cite{kool2019attention} solved vehicle routing and TSP instances of varying size with a single trained model; similar ideas appear in job-shop scheduling \cite{park2021jobshop}. Self-attention handles variable cardinality because the softmax normalizes over whatever set is present.

Training on small instances and deploying on larger ones—scale generalization—has been explored through curriculum learning and size-invariant architectural biases \cite{bengio2021ml4co}, but remains open. We combine cross-attention between a single task query and a variable-size server key set with explicit feature regularization to achieve zero-shot transfer across cluster sizes.

\section{Problem Formulation} \label{Problem Formulation}

We consider online scheduling of agentic workflows on a heterogeneous computing cluster. Workflows arrive over time, each composed of interdependent primitives. The scheduler assigns ready primitives to servers in real time, aiming to minimize average response time subject to resource constraints and dependency ordering.

\subsection{Workflow Model}

Let \(\mathcal{W} = \{\mathcal{W}_1,\mathcal{W}_2,\dots\}\) denote the set of workflows that arrive during the system's operation. Each workflow \(\mathcal{W}_k\) is represented as a directed acyclic graph (DAG) \(\mathcal{G}_k = (\mathcal{V}_k,\mathcal{E}_k)\), where:
\begin{itemize}
    \item \(\mathcal{V}_k\) is the set of \textit{primitives} (atomic execution units). Each primitive \(v_i \in \mathcal{V}_k\) is characterized by a tuple \((c_i, m_i, d_i)\):
        \begin{itemize}
            \item \(c_i > 0\): computational demand (in GFLOPS),
            \item \(m_i > 0\): memory requirement (in GB),
            \item \(d_i \geq 0\): output data volume (in GB) to be transmitted to successor tasks.
        \end{itemize}
    \item \(\mathcal{E}_k \subseteq \mathcal{V}_k \times \mathcal{V}_k\) is the set of directed edges representing data dependencies. An edge \((v_i, v_j) \in \mathcal{E}_k\) indicates that \(v_j\) depends on the completion of \(v_i\).
\end{itemize}

The \textit{predecessor set} of a node \(v_j\) is \(\Gamma^{-}(v_j) = \{v_i \in \mathcal{V}_k : (v_i, v_j) \in \mathcal{E}_k\}\). Node \(v_j\) becomes \textit{ready} when all its predecessors have finished execution, i.e., \(\Gamma^{-}(v_j) \subseteq \mathcal{D}(t)\), where \(\mathcal{D}(t)\) is the set of tasks completed by time \(t\). The \textit{ready set} at time \(t\) is
\[
\mathcal{R}(t) = \left\{ v_j \in \bigcup_k \mathcal{V}_k \setminus \mathcal{D}(t) \;:\; \Gamma^{-}(v_j) \subseteq \mathcal{D}(t) \right\},
\]
which constitutes the only schedulable tasks at that moment. The composition of \(\mathcal{R}(t)\) changes dynamically as tasks complete and new workflows arrive. Workflows arrive according to a Poisson process; the arrival rate may vary with system load.

\subsection{Cluster Model}

The computing cluster consists of a set of servers \(\mathcal{H} = \{h_1, \dots, h_N\}\), where \(N\) can change over time due to auto-scaling, preemption, or failures. Each server \(h_j\) is characterized by its resource capacities \((C_j, M_j)\):
\begin{itemize}
    \item \(C_j > 0\): computing capacity (in GFLOPS),
    \item \(M_j > 0\): memory capacity (in GB).
\end{itemize}
Servers are heterogeneous; capacities may differ by orders of magnitude.

Each server executes at most one primitive at a time. A server \(h_j\) that is idle and satisfies the memory constraint \(m_i \leq M_j\) may accept the assignment of a task \(v_i\). Once assigned, the \textit{execution time} of \(v_i\) on \(h_j\) is
\[
\tau_{ij} = \frac{c_i}{C_j}.
\]

If a dependency edge \((v_i, v_j) \in \mathcal{E}_k\) has its predecessor \(v_i\) assigned to server \(h_p\) and its successor \(v_j\) assigned to a different server \(h_q\) (\(p \neq q\)), the cross-server data transfer incurs a \textit{communication cost}
\[
\delta_{ij} = \frac{d_i}{B} + L,
\]
where \(B\) is the cluster network bandwidth (GB/s) and \(L\) is the base network latency (s). If \(p = q\), then \(\delta_{ij} = 0\). This communication cost does not directly affect the completion time of a task; instead, it serves as a reward-shaping signal to encourage locality (see Section~\ref{sec:reward}).

\subsection{Objective and Constraints}

Let \(f_i\) denote the actual completion time of task \(v_i\) (including queuing and execution delays). Let \(a_{k(i)}\) be the arrival time of the workflow \(\mathcal{W}_{k(i)}\) to which \(v_i\) belongs. The \textit{response time} of \(v_i\) is \(T_i = f_i - a_{k(i)}\). The scheduler aims to find a scheduling policy $\pi$ to minimize the \textit{average response time} over all completed tasks:
\[
\min \; \bar{T} = \frac{1}{|\mathcal{P}|} \sum_{v_i \in \mathcal{P}} \left( f_i - a_{k(i)} \right),
\]
where \(\mathcal{P}\) is the set of all completed tasks during the operating period.

The schedule must satisfy the following constraints at all times:
\begin{itemize}
    \item \textbf{Readyness:} A task can only be assigned when it is in the ready set \(\mathcal{R}(t)\).
    \item \textbf{Server idleness:} A server can execute at most one task at a time; it must be idle at the moment of assignment.
    \item \textbf{Memory capacity:} The assigned task's memory requirement must not exceed the server's memory capacity: \(m_i \leq M_j\).
    \item \textbf{Non-preemption:} Once a task starts on a server, it runs to completion without interruption.
\end{itemize}

However, this formulation is incomplete without specifying how the policy $\pi$ is structured. We defer the full MDP definition—including state, action, and reward—to Section~\ref{RL Formulation}, after introducing the RL framework.

\subsection{Challenges and MDP Formulation}

Compared to classical static DAG scheduling, this problem introduces additional difficulties:
\begin{itemize}
    \item \textbf{Online arrival:} Workflows arrive stochastically; the scheduler has no knowledge of future tasks.
    \item \textbf{DAG dependencies:} Readiness depends on predecessor completions, and cross-server placement incurs communication costs, so greedy assignment is generally suboptimal.
    \item \textbf{Variable cluster size:} $N$ may change dynamically through auto-scaling, making it necessary for the scheduler to work at scales it was never trained on.
    \item \textbf{Large action space:} The number of feasible task-server pairings grows combinatorially with the ready set and the server count.
\end{itemize}

We model the problem as a Markov Decision Process (MDP). The state encodes the current task (chosen by a deterministic first-level heuristic) together with all server statuses. The action picks a server for that task. The reward encourages short execution times and data locality, aligning with the response-time objective. The full MDP specification appears in Section~\ref{Algorithm}.

\section{Algorithm} \label{Algorithm}

\subsection{Reinforcement Learning Formulation} \label{RL Formulation}

We define the MDP $(\mathcal{S}, \mathcal{A}, r, \mathcal{T}, \gamma)$ as follows. Letting the agent jointly decide over all ready tasks and all servers would produce a Cartesian-product action space that grows too quickly. We therefore split scheduling into two levels. The first level is deterministic: it selects a single task $v^*$ from $\mathcal{R}(t)$ by longest hop-count path (in edges) to the terminal node of its workflow, prioritizing critical-path tasks. No parameters are learned at this level. The second level is the RL agent: given $v^*$ and the full cluster state, it chooses which server to assign $v^*$ to. This decomposition reduces each decision step to a single server selection.

The formal optimization objective is
\begin{subequations} \label{eq:optimization_objective}
\begin{align}
\min_{\pi:\,\mathcal{S}\to\mathcal{A}} \quad & \bar{T} = \frac{1}{|\mathcal{P}|}\sum_{v_i \in \mathcal{P}} \bigl(f_i - a_{k(i)}\bigr), \\
\mathrm{s.t.} \quad & a_t = \pi(\boldsymbol{s}_t), \quad \forall\, t, \\
& v^* \in \mathcal{R}(t), \\
& h_{a_t} \in \mathcal{I}(t), \\
& m_{v^*} \leq M_{a_t},
\end{align}
\end{subequations}
where $\boldsymbol{s}_t \in \mathcal{S}$ is the system state at decision step $t$, $a_t \in \mathcal{A}$ is the server index selected by the policy, $v^*$ is the task chosen by the first-level scheduler, and $\mathcal{I}(t) \subseteq \mathcal{H}$ is the set of idle servers at time $t$. $\mathcal{P}$ denotes the set of all completed tasks during the operating period and $\bar{T}$ is the mean response time. The constraint $a_t = \pi(\boldsymbol{s}_t)$ makes explicit that every server assignment is determined by the same stationary policy acting on the current state, so optimizing $\pi$ over the state space $\mathcal{S}$ is the sole degree of freedom for minimizing $\bar{T}$.

\subsubsection{State $\mathcal{S}$}
At each decision step, the agent's state $\boldsymbol{s}_t$ is formed by concatenating the features of the current task $v^*$ with the real-time states of all $N$ servers. The task feature vector $\boldsymbol{\phi}_t \in \mathbb{R}^5$ contains five scalars: normalized computational demand $c_{v^*}/C_{\max}$, normalized memory requirement $m_{v^*}/M_{\max}$, normalized number of successors $|\Gamma^+(v^*)|/5$, the completion ratio of the parent workflow, and the output data volume $d_{v^*}$ (unnormalized). The server feature vector $\boldsymbol{\psi}_t^{(j)} \in \mathbb{R}^5$ for server $h_j$ contains its normalized computing capacity $C_j/C_{\max}$, normalized memory $M_j/M_{\max}$, remaining execution time of the current task, instantaneous power consumption, and the estimated communication delay from the server hosting a predecessor of $v^*$ to $h_j$ if $v^*$ were assigned there. The state vector is
\begin{equation*}
\boldsymbol{s}_t = \bigl[\boldsymbol{\phi}_t;\; \boldsymbol{\psi}_t^{(1)},\; \ldots,\; \boldsymbol{\psi}_t^{(N)}\bigr] \in \mathbb{R}^{5 + 5N},
\end{equation*}
whose dimensionality scales linearly with the cluster size $N$. Note that this state vector is only a conceptual description; in our cross-attention architecture, $\boldsymbol{\phi}_t$ and $\boldsymbol{\psi}_t^{(j)}$ are embedded separately.

\subsubsection{Action $\mathcal{A}$}
At each step, the agent selects an action $a_t$ from $\{1, \ldots, N\}$, indicating the assignment of $v^*$ to server $h_{a_t}$. If $h_{a_t}$ is currently idle and satisfies $m_{v^*} \leq M_{a_t}$, the assignment takes effect immediately and the simulation clock does not advance; the scheduler then selects the next task from the ready set. Otherwise, the action is invalid, and the simulation clock advances by $\Delta t$, during which progress updates for executing tasks, readiness releases upon task completions, and Poisson arrivals of new workflows are processed.

\subsubsection{Reward $R$} \label{sec:reward}
The immediate reward for a valid assignment is
\begin{equation} \label{eq:reward}
r_t = w_\tau \, e^{-\tau_{v^*, a_t}/s_\tau} - \alpha\,\frac{\delta_{v^*, a_t}}{\delta_{\max}} - \beta\,\frac{e_{v^*, a_t}}{e_{\max}},
\end{equation}
where $\tau_{v^*, a_t} = c_{v^*}/C_{a_t}$ is the estimated execution time, $\delta_{v^*, a_t}$ is the communication cost from the server hosting one of $v^*$'s already-assigned predecessors to $h_{a_t}$ (when $v^*$ has multiple predecessors, we use the first predecessor whose assignment is completed; if $v^*$ has no predecessors, $\delta_{v^*, a_t} = 0$), and $e_{v^*, a_t}$ is the estimated energy consumption. The normalization baselines are $\delta_{\max}$ and $e_{\max}$. We set $w_\tau = 1.0$, $s_\tau = 1.0$, $\alpha = 0.1$, and $\beta = 0$ (energy optimization is not active in the current experiments). Invalid actions receive $r_t = -0.001$ as a mild penalty to discourage their abuse. The exponential form bounds the time reward within $(0, 1]$, giving stronger marginal incentive for faster assignments while keeping the reward magnitude stable across tasks with different compute demands. The learning objective is $\max \mathbb{E}[\sum_{t=0}^{\infty} \gamma^t r_t]$ with $\gamma = 0.99$. Because the dominant term in \eqref{eq:reward} decreases monotonically with execution time, maximizing cumulative reward is aligned with minimizing $\bar{T}$ in \eqref{eq:optimization_objective}—though not strictly equivalent, since the communication and energy penalties also encourage locality and efficiency. One subtlety: the simulation clock advances only on invalid actions, so policies with higher invalid-action rates accumulate more workflow arrivals within the same simulated duration, inflating their completed-task count. We therefore use $\bar{T}$ as the evaluation metric, which is unaffected by this artifact.

\subsection{Cross-Attention Pointer Network} \label{Cross-Attention Pointer Network}

Given this MDP, the agent must pick a server at each step. We adopt a cross-attention pointer network: task features form the query, server features form the key-value set. After $L$ cross-attention layers, the task representation encodes information about the full cluster, and pointer-style dot-product scores yield the selection distribution over servers.

\subsubsection{Embedding}
We parse $\boldsymbol{s}_t$ into the task feature $\boldsymbol{\phi} \in \mathbb{R}^{d_0}$ ($d_0 = 5$) and the server feature matrix $\boldsymbol{\Psi} = [\boldsymbol{\psi}^{(1)}; \ldots; \boldsymbol{\psi}^{(N)}] \in \mathbb{R}^{N \times d_0}$. Two separate two-layer fully connected networks with ReLU activations map these into a shared $d$-dimensional latent space: $\boldsymbol{q}^{(0)} = g_p(\boldsymbol{\phi}) \in \mathbb{R}^d$ for the task and $\boldsymbol{z}_j = g_s(\boldsymbol{\psi}^{(j)}) \in \mathbb{R}^d$ for each server. The two networks $g_p: \mathbb{R}^{d_0} \to \mathbb{R}^d$ and $g_s: \mathbb{R}^{d_0} \to \mathbb{R}^d$ have independent parameters.

\subsubsection{Cross-Attention Backbone}
The task embedding $\boldsymbol{q}^{(0)}$ passes through $L$ successive cross-attention blocks. Each block has two sublayers—multi-head cross-attention followed by a feedforward network—both with residual connections and layer normalization. In layer $\ell$, the task embedding is the query and all server embeddings are key-value pairs, processed in parallel across $H$ heads. For head $h$ ($h = 1, \ldots, H$), separate projection matrices map query and key-values into a $d_h = d/H$ dimensional subspace. The scaled dot-product attention weights are
\begin{equation} \label{eq:cross_attention}
\alpha_{\ell j}^{(h)} = \frac{\exp\!\bigl((\boldsymbol{W}_{Qh}^{(\ell)}\boldsymbol{q}^{(\ell-1)})^\top (\boldsymbol{W}_{Kh}^{(\ell)}\boldsymbol{z}_j) \,/\, \sqrt{d_h}\bigr)}{\sum_{j'=1}^{N}\exp\!\bigl((\boldsymbol{W}_{Qh}^{(\ell)}\boldsymbol{q}^{(\ell-1)})^\top (\boldsymbol{W}_{Kh}^{(\ell)}\boldsymbol{z}_{j'}) \,/\, \sqrt{d_h}\bigr)}.
\end{equation}
Each head aggregates value vectors $\boldsymbol{W}_{Vh}^{(\ell)}\boldsymbol{z}_j$ weighted by these coefficients. The concatenated head outputs pass through a projection $\boldsymbol{W}_O^{(\ell)}$, then a residual connection and layer normalization yield the intermediate representation $\hat{\boldsymbol{q}}^{(\ell)}$. A feedforward sublayer—two linear layers with hidden dimension $4d$ and GELU activation, again with residual connection and layer normalization—produces the layer output $\boldsymbol{q}^{(\ell)}$. After $L$ layers,
\begin{equation*}
\boldsymbol{f} = \boldsymbol{q}^{(L)} \in \mathbb{R}^d,
\end{equation*}
is called the attention feature. It compresses the contextual information of all $N$ servers into a fixed-length vector whose dimension $d$ is independent of the cluster size $N$.

\subsubsection{Actor and Critic}
The Actor uses $\boldsymbol{f}$ as query and the server embeddings $\{\boldsymbol{z}_j\}$ as keys, producing selection scores via pointer-style dot products:
\begin{equation} \label{eq:pointer}
p_j = (\boldsymbol{W}_q\,\boldsymbol{f})^\top(\boldsymbol{W}_k\,\boldsymbol{z}_j), \quad \pi_\theta(a_t = j \mid \boldsymbol{s}_t) = \frac{e^{p_j}}{\sum_{j'=1}^{N} e^{p_{j'}}},
\end{equation}
where $\boldsymbol{W}_q, \boldsymbol{W}_k \in \mathbb{R}^{d \times d}$ are learnable projections. The Actor's output dimension equals the current cluster size $N$ and adapts at test time. The Critic takes $\boldsymbol{f}$ alone and estimates $V_\theta(\boldsymbol{s}_t)$ through a three-layer MLP (hidden dimension 64, ReLU). It deliberately avoids aggregating over $\{\boldsymbol{z}_j\}$ so as not to introduce an implicit dependency on $N$. The full network is trained end-to-end with PPO.

\subsubsection{Permutation Invariance and Scale Adaptability}
No positional encoding is used, so the ordering of server embeddings $\{\boldsymbol{z}_j\}$ does not affect $\boldsymbol{f}$—the network is permutation-invariant over the server set. The Actor's output dimension equals $N$ automatically. These two properties allow the model to accept any cluster size at test time. Permutation invariance, however, is necessary but not sufficient for generalization. When $N' \neq N_{\mathrm{train}}$, we observe that the distribution of $\boldsymbol{f}$ shifts systematically and decision quality degrades. Controlling the feature statistics is therefore required on top of the architectural design.

\begin{algorithm}[t]
\caption{Cross-Attention Pointer Network + SRR Training}
\label{algorithm:training}
\begin{algorithmic}[1]
\Require Environment $\mathcal{E}$ ($N$ servers), network parameters $\theta$, PPO hyperparameters $(\gamma, \lambda_{\mathrm{GAE}}, \varepsilon, c_1, c_2)$, SRR weights $(\lambda_1, \lambda_2)$, warmup start epoch $E_s$, warmup duration $E_w$, total training epochs $E$, collection steps per epoch $T$, update epochs $K$, mini-batch size $B$
\Ensure Trained policy parameters $\theta^*$
\State Randomly initialize $\theta$; initialize optimizer and cosine annealing learning rate scheduler
\For{$e = 1$ to $E$}
    \State Clear experience buffer $\mathcal{D} \leftarrow \emptyset$
    \For{$t = 1$ to $T$}
        \State Obtain state $\boldsymbol{s}_t$ from $\mathcal{E}$; parse into $(\boldsymbol{\phi},\, \boldsymbol{\Psi})$
        \State $\boldsymbol{q}^{(0)} \leftarrow g_p(\boldsymbol{\phi})$, $\boldsymbol{z}_j \leftarrow g_s(\boldsymbol{\psi}^{(j)})$, $j = 1, \ldots, N$
        \For{$\ell = 1$ to $L$}
            \State $\boldsymbol{q}^{(\ell)} \leftarrow \mathrm{CrossAttnBlock}(\boldsymbol{q}^{(\ell-1)}, \{\boldsymbol{z}_j\})$
        \EndFor
        \State $\boldsymbol{f} \leftarrow \boldsymbol{q}^{(L)}$
        \State $p_j \leftarrow (\boldsymbol{W}_q \boldsymbol{f})^\top (\boldsymbol{W}_k \boldsymbol{z}_j)$, $\pi_\theta(j \mid \boldsymbol{s}_t) \leftarrow e^{p_j} / \sum_{j'} e^{p_{j'}}$
        \State Sample $a_t$ from $\pi_\theta$; compute $\log \pi_\theta(a_t \mid \boldsymbol{s}_t)$ and $V_\theta(\boldsymbol{s}_t) \leftarrow \mathrm{MLP}(\boldsymbol{f})$
        \State Execute $a_t$ in $\mathcal{E}$; receive $r_t$, $\boldsymbol{s}_{t+1}$
        \State $\mathcal{D} \leftarrow \mathcal{D} \cup \{(\boldsymbol{s}_t, a_t, r_t, \log \pi_\theta(a_t \mid \boldsymbol{s}_t), V_\theta(\boldsymbol{s}_t))\}$
    \EndFor
    \State Compute advantages $\hat{A}_t$ and returns $R_t$ using GAE, $\forall\, (\cdot) \in \mathcal{D}$
    \State $w \leftarrow \min\!\bigl(1,\, \max(0,\, (e - E_s) / E_w)\bigr)$
    \For{$k = 1$ to $K$}
        \For{each mini-batch $\mathcal{B} \subset \mathcal{D}$, $|\mathcal{B}| = B$}
            \State Recompute forward pass to obtain $\log \pi_\theta'$, entropy $H$, $V_\theta'$, $\boldsymbol{f}$
            \State $\rho \leftarrow \exp(\log \pi_\theta' - \log \pi_{\theta_{\mathrm{old}}})$
            \State $\mathcal{L}_{\mathrm{clip}} \leftarrow -\frac{1}{|\mathcal{B}|}\sum \min\!\bigl(\rho\,\hat{A},\, \mathrm{clip}(\rho,\, 1{-}\varepsilon,\, 1{+}\varepsilon)\,\hat{A}\bigr)$
            \State $\mathcal{L}_{\mathrm{vf}} \leftarrow \frac{1}{|\mathcal{B}|}\sum (R - V_\theta')^2$
            \State $\mathcal{L}_{\pi} \leftarrow \mathcal{L}_{\mathrm{clip}} + c_1\,\mathcal{L}_{\mathrm{vf}} - c_2\,H$
            \State Compute $\hat{\boldsymbol{C}}$ from $\boldsymbol{f}$ in $\mathcal{B}$; $\mathcal{L}_{c} \leftarrow \sum_{i \neq j} |(\hat{\boldsymbol{C}})_{ij}|$
            \State Compute per-dimension $\hat{\mu}_k$, $\hat{\sigma}_k^2$; $\mathcal{L}_{d} \leftarrow \frac{1}{2d}\sum_k (\hat{\mu}_k^2 + \hat{\sigma}_k^2 - \ln\hat{\sigma}_k^2 - 1)$
            \State $\mathcal{L} \leftarrow \mathcal{L}_{\pi} + w\,(\lambda_1\,\mathcal{L}_{c} + \lambda_2\,\mathcal{L}_{d})$ \hfill (see \eqref{eq:total_loss})
            \State $\theta \leftarrow \theta - \eta\,\nabla_\theta \mathcal{L}$ (update after gradient clipping)
        \EndFor
    \EndFor
    \State Update learning rate scheduler
\EndFor
\State \Return $\theta^* \leftarrow \theta$
\end{algorithmic}
\end{algorithm}

\subsection{Structured Representation Regularization} \label{SRR}

Cross-attention with a pointer network provides permutation invariance and variable output size, but does nothing to prevent the attention feature $\boldsymbol{f}$ from drifting when the cluster grows. This drift manifests in two ways. When dimensions of $\boldsymbol{f}$ are correlated, effective information concentrates along a few principal directions; a change in $N'$ shifts those directions and the Actor loses its ability to discriminate among servers. Separately, the softmax denominator $\sum_{j'} e^{p_{j'}}$ in \eqref{eq:pointer} grows with $N'$, flattening selection probabilities and blurring the distinction between good and bad servers.

We introduce Structured Representation Regularization (SRR) to counter both effects. SRR imposes two constraints on the batch statistics of $\boldsymbol{f}$ during training. A decorrelation loss $\mathcal{L}_{c}$ penalizes off-diagonal entries of the empirical covariance matrix, pushing feature dimensions toward independence so that information distributes across all $d$ dimensions rather than concentrating on a few axes. A distribution constraint $\mathcal{L}_{d}$ pulls each dimension's marginal toward $\mathcal{N}(0,1)$ via a KL penalty, anchoring the statistics to a fixed reference irrespective of how many servers participate in the attention computation. The combined effect is that features maintain stable structure as $N$ varies, producing more confident policy outputs that resist the probability-flattening caused by a larger softmax denominator. Below we detail the two loss terms.

Let the set of attention features in the current mini-batch be $\boldsymbol{F} = [\boldsymbol{f}_1, \ldots, \boldsymbol{f}_B]^\top \in \mathbb{R}^{B \times d}$, with batch mean $\bar{\boldsymbol{f}} = \frac{1}{B}\sum_{b} \boldsymbol{f}_b$ and empirical covariance matrix
\begin{equation*}
\hat{\boldsymbol{C}} = \frac{1}{B-1}\bigl(\boldsymbol{F} - \boldsymbol{1}\bar{\boldsymbol{f}}^\top\bigr)^\top\!\bigl(\boldsymbol{F} - \boldsymbol{1}\bar{\boldsymbol{f}}^\top\bigr) \in \mathbb{R}^{d \times d}.
\end{equation*}

The decorrelation loss penalizes the off-diagonal elements of $\hat{\boldsymbol{C}}$, pushing the dimensions of $\boldsymbol{f}$ toward statistical independence:
\begin{equation*}
\mathcal{L}_{c} = \sum_{i \neq j} \bigl|(\hat{\boldsymbol{C}})_{ij}\bigr|.
\end{equation*}

\begin{algorithm}[t]
\caption{Online Scheduling Decision}
\label{algorithm:decision}
\begin{algorithmic}[1]
\Require Trained parameters $\theta^*$, target cluster $\mathcal{H}' = \{h_1, \ldots, h_{N'}\}$ ($N'$ may differ from $N$)
\Ensure Assignment decision mapping each ready task $v^*$ to a server
\State Load $\theta^*$; set to evaluation mode
\While{system is running}
    \State Wait until ready set $\mathcal{R}(t) \neq \emptyset$
    \State First-level scheduler selects $v^*$ from $\mathcal{R}(t)$ (prioritizing the longest hop count to the workflow's terminal node)
    \State Construct state $\boldsymbol{s}_t = [\boldsymbol{\phi};\, \boldsymbol{\psi}^{(1)}, \ldots, \boldsymbol{\psi}^{(N')}]$
    \State $\boldsymbol{q}^{(0)} \leftarrow g_p(\boldsymbol{\phi})$, $\boldsymbol{z}_j \leftarrow g_s(\boldsymbol{\psi}^{(j)})$, $j = 1, \ldots, N'$
    \For{$\ell = 1$ to $L$}
        \State $\boldsymbol{q}^{(\ell)} \leftarrow \mathrm{CrossAttnBlock}(\boldsymbol{q}^{(\ell-1)}, \{\boldsymbol{z}_j\}_{j=1}^{N'})$
    \EndFor
    \State $\boldsymbol{f} \leftarrow \boldsymbol{q}^{(L)}$
    \State $a^* \leftarrow \arg\max_{j \in \{1,\ldots,N'\}} (\boldsymbol{W}_q \boldsymbol{f})^\top (\boldsymbol{W}_k \boldsymbol{z}_j)$
    \If{$h_{a^*}$ is idle \textbf{and} $m_{v^*} \leq M_{a^*}$}
        \State Assign $v^*$ to $h_{a^*}$; execute immediately
    \Else
        \State Mark as invalid action; advance simulation clock by $\Delta t$
    \EndIf
\EndWhile
\end{algorithmic}
\end{algorithm}

The distribution constraint loss drives the marginal distribution of each dimension toward the standard normal, equivalent to minimizing the sum of per-dimension KL divergences from $\mathcal{N}(0,1)$:
\begin{equation*}
\mathcal{L}_{d} = D_{KL}\bigl(p(\boldsymbol{f}) \,\|\, \mathcal{N}(\boldsymbol{0}, \boldsymbol{I})\bigr) = \frac{1}{2d}\sum_{k=1}^{d}\bigl(\hat{\mu}_k^2 + \hat{\sigma}_k^2 - \ln\hat{\sigma}_k^2 - 1\bigr),
\end{equation*}
where $\hat{\mu}_k = \frac{1}{B}\sum_{b} f_{bk}$ and $\hat{\sigma}_k^2 = \frac{1}{B-1}\sum_{b} (f_{bk} - \hat{\mu}_k)^2$. We apply both constraints to $\boldsymbol{f}$ rather than to the embedding layer $\{\boldsymbol{z}_j\}$. The rationale is that $\boldsymbol{f}$ feeds directly into Actor scoring and Critic estimation; stabilizing its distribution stabilizes the entire decision pipeline. The embedding layer, by contrast, operates on each server independently and its statistics do not inherently depend on $N$. The total training loss is
\begin{equation} \label{eq:total_loss}
\mathcal{L} = \mathcal{L}_{\pi} + \lambda_1\,\mathcal{L}_{c} + \lambda_2\,\mathcal{L}_{d},
\end{equation}
with $\lambda_1 = 0.01$ and $\lambda_2 = 0.001$. SRR introduces no additional network parameters. To avoid disrupting early-stage policy learning while representations are still forming, we linearly ramp $\lambda_1$ and $\lambda_2$ from zero starting at epoch 10, reaching full strength by epoch 60.

Once the constraints are approximately satisfied—$\mathbb{E}[\boldsymbol{f}] \approx \boldsymbol{0}$ and $\hat{\boldsymbol{C}} \approx \boldsymbol{I}_d$—each dimension of $\boldsymbol{W}_q \boldsymbol{f}$ contributes independently with stable magnitude. A larger $N$ changes the aggregation pattern but no longer concentrates information along shifted principal directions. SRR thus converts the architectural property of permutation invariance into actual runtime robustness across scales.

\section{Experiment} \label{Experiment And Result}

\subsection{Experimental Setup}

\subsubsection{Hardware}
All simulation experiments are conducted on a device equipped with an Apple M4 processor and 32\,GB of memory.

\subsubsection{Simulation Setup}
The simulated cluster has three tiers of servers at a base scale of $N{=}16$ (Table~\ref{tab:server_config}). For generalization tests, we scale to $N{=}32$ and $N{=}48$ by proportionally increasing each server type, keeping the small:medium:large ratio at roughly 2:5:1. Workflows arrive as a Poisson process with base rate $\lambda = 1.0$; the rate scales with cluster size ($\lambda{=}2.0$ at $N{=}32$, $\lambda{=}3.0$ at $N{=}48$) to maintain comparable load pressure. Each workflow is a randomly generated DAG containing 3--8 primitives (uniform). Computation demands are drawn from $\mathcal{N}(10.0, 3.0^2)$ truncated below at 1.0. Memory demands follow $\mathcal{N}(16.0, 8.0^2)$ clipped to $[1.0, 80.0]$. Inter-task data volumes follow $\mathcal{N}(0.5, 0.2^2)$\,GB with a floor of 0.01\,GB. Cross-server communication latency equals data size divided by bandwidth (10\,GB/s) plus 0.01\,s base latency; intra-server transfers are free.

\begin{table}[t]
\caption{Heterogeneous Server Configuration}
\label{tab:server_config}
\centering
\begin{tabular}{lccc}
\toprule
 & {Small} & {Medium} & {Large} \\
\midrule
Compute     & 1.0 & 2.0 & 4.0 \\
Memory (GB) & 8   & 32  & 80  \\
Idle Power (W)  & 30  & 50  & 80  \\
Busy Power (W)  & 100 & 200 & 400 \\
\midrule
Count ($N{=}16$) & 4  & 10 & 2  \\
Count ($N{=}32$) & 8  & 20 & 4  \\
Count ($N{=}48$) & 12 & 30 & 6  \\
\bottomrule
\end{tabular}
\end{table}

\subsubsection{Training and Evaluation}
All models are trained on the $N{=}16$ cluster using PPO. Table~\ref{tab:hyperparams} summarizes the key hyperparameters. For the cross-attention architecture, the embedding dimension is 16, with 4 attention heads and 2 cross-attention layers. The SRR regularization activates at epoch 10 with linear warmup over 50 epochs. Four parallel environments are used for data collection with a fixed random seed of 42. Each configuration is evaluated over 30 episodes with fixed random seeds, reporting the mean and standard deviation of the average response time as the primary metric.

\begin{table}[t]
\caption{Training Hyperparameters}
\label{tab:hyperparams}
\centering
\begin{tabular}{ll}
\toprule
{Parameter} & {Value} \\
\midrule
Learning rate        & $3 \times 10^{-4} \to 5 \times 10^{-5}$, cosine decay \\
Discount $\gamma$    & 0.99 \\
GAE $\lambda$        & 0.95 \\
Clip $\epsilon$      & 0.2 \\
Entropy coeff.       & 0.01 \\
Grad clip            & 0.5 \\
Epochs               & 1000 \\
Steps / collect      & 1024 \\
Batch size           & 256 \\
Updates / collect    & 8 \\
Hidden layers        & $[256, 256]$ \\
SRR $\lambda_{\text{cov}}$ & 0.01 \\
SRR $\lambda_{\text{kl}}$  & 0.001 \\
\bottomrule
\end{tabular}
\end{table}

\subsection{Baselines}

We compare SCALE against five baselines:

\begin{itemize}
    \item \textbf{CrossAttn}: The same cross-attention pointer network but without SRR, isolating the regularization's contribution.
    \item \textbf{PPO}: Standard PPO with a structured MLP policy that processes primitive and server features separately before concatenation. Output dimension is fixed at $N{=}16$; it cannot run on larger clusters.
    \item \textbf{PPO+SRR}: PPO augmented with the SRR loss. The output dimension remains fixed, so it still cannot scale, but it tests whether SRR helps a non-scalable architecture.
    \item \textbf{DQN}: Deep Q-Network with $\epsilon$-greedy exploration and dueling architecture.
    \item \textbf{SAC}: Soft Actor-Critic with entropy-regularized objective.
\end{itemize}

MLP-based methods (PPO, PPO+SRR, DQN, SAC) have fixed output dimensionality and can only run at $N{=}16$. Cross-attention methods support zero-shot deployment at arbitrary cluster sizes.

\subsection{Main Results}

\begin{table}[t]
\caption{16-Node Performance Comparison}
\label{tab:16node_results}
\centering
\begin{tabular}{lcc}
\toprule
{Method} & {Completed Tasks} & {Resp. Time (s)} \\
\midrule
{SCALE}  & $8.37 \pm 2.15$ & $4.58 \pm 0.68$ \\
CrossAttn        & $8.10 \pm 1.85$ & $4.65 \pm 0.67$ \\
SAC              & $5.93 \pm 1.41$ & $4.50 \pm 0.29$ \\
PPO              & $4.03 \pm 0.71$ & $4.37 \pm 0.73$ \\
PPO+SRR          & $4.30 \pm 1.10$ & $4.45 \pm 0.73$ \\
DQN              & $4.00 \pm 0.68$ & $4.88 \pm 0.49$ \\
\bottomrule
\end{tabular}
\end{table}

Table~\ref{tab:16node_results} compares all six methods at $N{=}16$. SCALE completes the most tasks per episode (8.37) at a response time of 4.58\,s; CrossAttn follows closely with 8.10 tasks. SAC lands in between at 5.93 tasks. PPO, PPO+SRR, and DQN all hover around 4 tasks per episode, with DQN showing the worst response time (4.88\,s). Adding SRR to PPO slightly improves throughput (4.03 $\to$ 4.30) without degrading response time, indicating that the regularization provides some benefit even on architectures that lack scale flexibility.

\begin{figure}[t]
\centering
\includegraphics[width=\columnwidth]{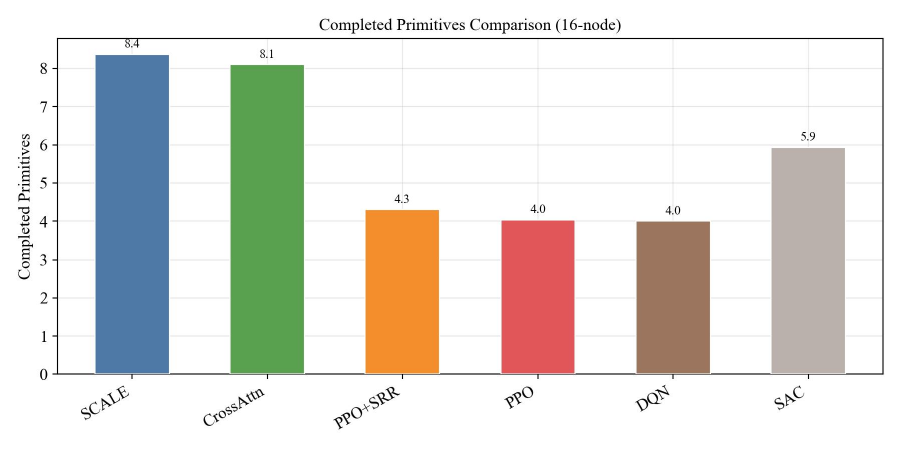}
\caption{16-Node Completed primitives comparison.}
\label{fig:16node_primitives}
\end{figure}

\begin{figure}[t]
\centering
\includegraphics[width=\columnwidth]{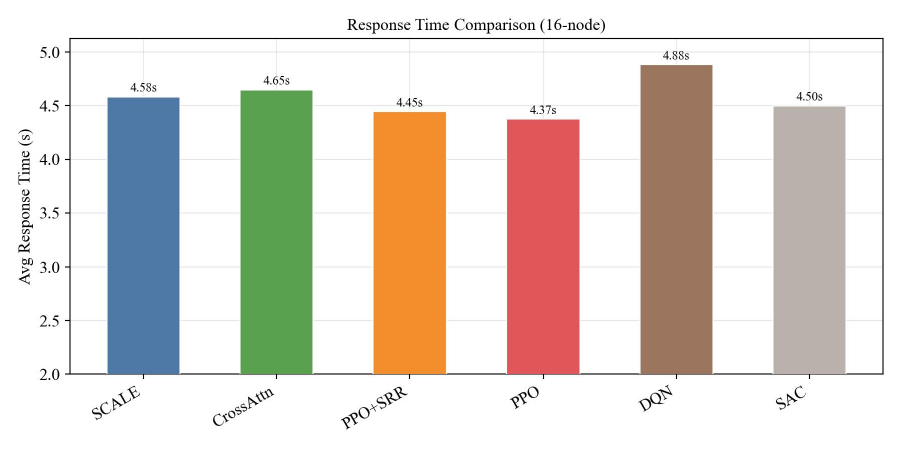}
\caption{16-Node Average response time comparison.}
\label{fig:16node_response_time}
\end{figure}

\subsection{Zero-Shot Scale Generalization}

\begin{table}[t]
\caption{Zero-Shot Generalization.}
\label{tab:generalization}
\centering
\begin{tabular}{lccc}
\toprule
{Method} & $N{=}16$ (train) & $N{=}32$ & $N{=}48$ \\
\midrule
{SCALE} & $4.58 \pm 0.68$ & $5.57 \pm 0.65$ & $6.22 \pm 0.70$ \\
CrossAttn       & $4.65 \pm 0.67$ & $5.70 \pm 0.52$ & $6.83 \pm 0.63$ \\
\bottomrule
\end{tabular}
\end{table}

\begin{figure}[t]
\centering
\includegraphics[width=\columnwidth]{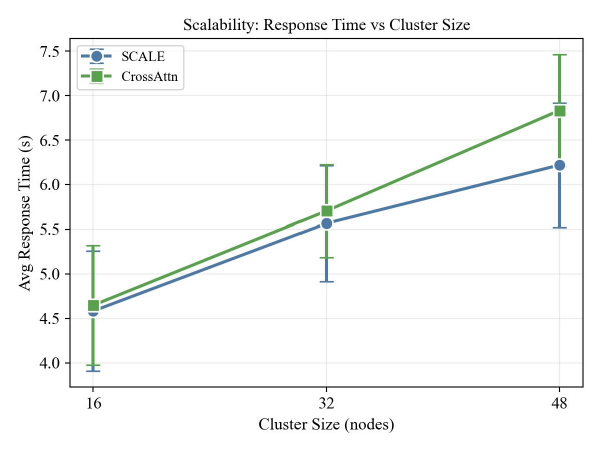}
\caption{Average response time across cluster scales. SCALE maintains stable response times as the cluster grows from 16 to 48 nodes, while CrossAttn degrades more rapidly.}
\label{fig:scale_response_time}
\end{figure}

Table~\ref{tab:generalization} reports response time for the two cross-attention variants across scales (MLP-based methods cannot run at $N \neq 16$). At the training scale, the two are nearly tied: 4.58\,s vs.\ 4.65\,s. A gap opens at $N{=}32$ (5.57 vs.\ 5.70) and widens at $N{=}48$, where SCALE achieves 6.22\,s against CrossAttn's 6.83\,s—an 8.9\% reduction. The pattern is consistent with feature drift accumulating as $N$ grows: without SRR, the pointer scores lose calibration; with SRR, they remain stable.

\subsection{Training Results}

\begin{figure}[t]
\centering
\includegraphics[width=\columnwidth]{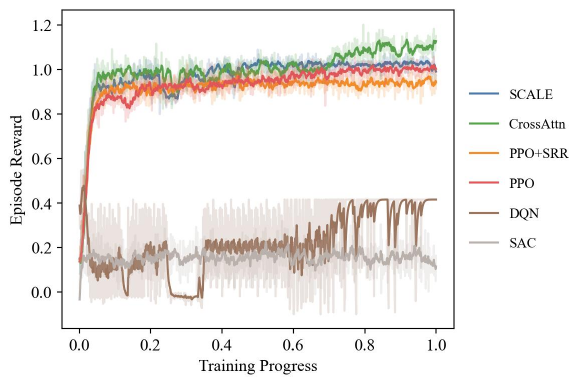}
\caption{Training reward curves of all methods.}
\label{fig:train_reward}
\end{figure}

Fig.~\ref{fig:train_reward} shows training curves. All methods rise quickly in the first 10\% of training. CrossAttn then continues climbing with large fluctuations, eventually exceeding 1.1—likely overfitting to the 16-node configuration. SCALE plateaus near 1.0 with much smaller variance, consistent with SRR constraining the representation space. PPO and PPO+SRR converge between 0.93 and 0.95, limited by MLP capacity. SAC converges slowly to about 0.6, and DQN is the slowest and most unstable, matching its weak evaluation performance.

\subsection{Attention Feature Quality Analysis}

\begin{table}[t]
\caption{Attention Feature Statistics}
\label{tab:feature_stats}
\centering
\begin{tabular}{llccc}
\toprule
{Metric} & {Method} & $N{=}16$ & $N{=}32$ & $N{=}48$ \\
\midrule
\multirow{2}{*}{Cov Off-Diag ($\downarrow$)}
    & {SCALE}   & {0.037} & {0.036} & {0.044} \\
    & CrossAttn         & 0.274          & 0.103          & 0.090 \\
\midrule
\multirow{2}{*}{KL to $\mathcal{N}(0,I)$ ($\downarrow$)}
    & {SCALE}   & 0.516 & 0.637 & 0.623 \\
    & CrossAttn         & 0.378 & 0.498 & 0.484 \\
\bottomrule
\end{tabular}
\end{table}

To see why SRR helps, we examine the statistics of the attention feature $\mathbf{z} \in \mathbb{R}^{d}$ directly.

Table~\ref{tab:feature_stats} reports two metrics across scales. Off-diagonal covariance measures inter-dimensional coupling. SCALE keeps this below 0.05 at all three scales; CrossAttn ranges from 0.09 to 0.27, indicating much stronger correlation among its feature dimensions. CrossAttn does have lower KL divergence at $N{=}16$—its features happen to look roughly Gaussian without being forced to—but this property breaks down at larger $N$, which is exactly the failure mode we target. What matters for generalization is decorrelation: when dimensions are independent, a change in $N$ perturbs each dimension locally instead of cascading through the entire vector.

\section{Discussion} \label{Discussion}

\subsection{Interpretation of Results}
SCALE generalizes from $N=16$ to $N=48$ without retraining, cutting average response time by 8.9\% relative to the unregularized baseline at the largest tested scale. At the training scale itself ($N=16$), SCALE does not outperform PPO or SAC on per-task latency—it completes more tasks per episode but at slightly higher per-task response time. This trade-off is expected: SRR constrains the representation in ways that sacrifice some in-distribution optimality for out-of-distribution robustness. The training curves corroborate this reading—CrossAttn without SRR reaches higher reward but with large variance, consistent with overfitting to the 16-node setting.

\subsection{Limitations}
Several limitations should be noted. Our evaluation reaches only $N=48$ (3$\times$ the training size); behavior at $N=128$ or beyond is untested. The generalization experiments hold the server-type ratio (Small:Medium:Large) constant, so a cluster with a different hardware mix could behave differently. The DAGs are randomly generated with limited topological variety—real agentic workflows may exhibit deeper chains or wider fan-outs. The first-level task selector uses a simple longest-path heuristic rather than a learned policy. All results are from simulation; deployment on a physical cluster running an actual agentic framework (e.g., LangGraph) would be needed to validate practical utility.

\subsection{Future Work}
The most immediate extension is testing at much larger scales ($N \geq 128$), where sparse attention approximations may become necessary for fast inference. Replacing the heuristic first-level scheduler with a learned joint task-and-server selection policy is another natural direction. Deployment inside a real agentic framework on physical hardware would clarify how well these simulation results transfer.

\section{Conclusion} \label{Conclusion}

We presented SCALE, a DRL scheduler for agentic workflow DAGs that trains once on a small cluster and deploys on larger ones without retraining. The architecture—cross-attention between a task query and a variable-size server key set, with pointer-network scoring—handles arbitrary $N$ by construction. Our main finding is that architectural flexibility alone is insufficient: without SRR, performance degrades as $N$ grows due to distribution shift in the attention feature. The decorrelation and KL-to-normal penalties in SRR stabilize these statistics, producing an 8.9\% response-time reduction over the unregularized baseline at $N{=}48$.

For elastic clusters where nodes are added or removed frequently, a scheduler requires both a size-agnostic architecture and explicit regularization of its internal representations. SCALE demonstrates this principle, though validation at much larger scales and on physical hardware remains to be done.

\bibliographystyle{elsarticle-num}
\bibliography{references}

\section*{Author Biographies}

\authorentry{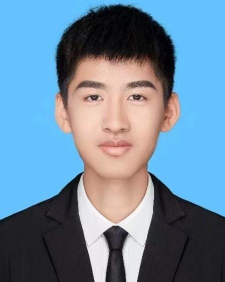}{Zhifei Xu}{is currently pursuing a bachelor's degree in the Faculty of Arts and Sciences, Beijing Normal University at Zhuhai, China. His main research interests include edge computing, reinforcement learning and their applications.}

\authorentry{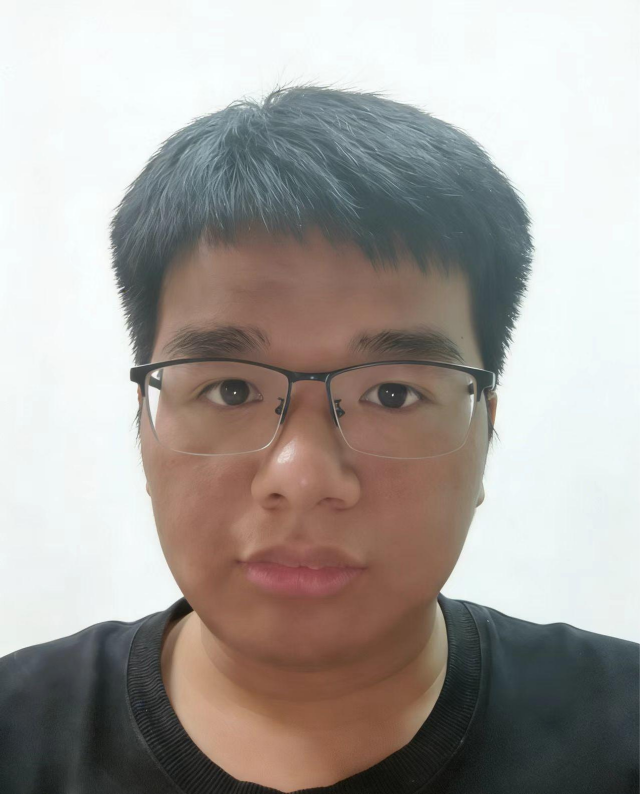}{Jierui Lan}{is currently pursuing a bachelor's degree in the Applied Statistics, Beijing Normal University at Zhuhai, China. His main research interests include representation learning, optimization algorithm, reinforcement learning and their applications.}

\authorentry{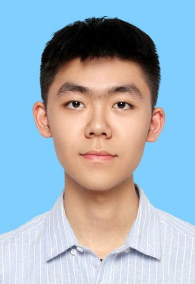}{Zixuan Liang}{is currently pursuing a bachelor's degree in the Applied Statistics, Beijing Normal University at Zhuhai, China. His main research interests include multiple testing, reinforcement learning and their applications.}

\authorentry{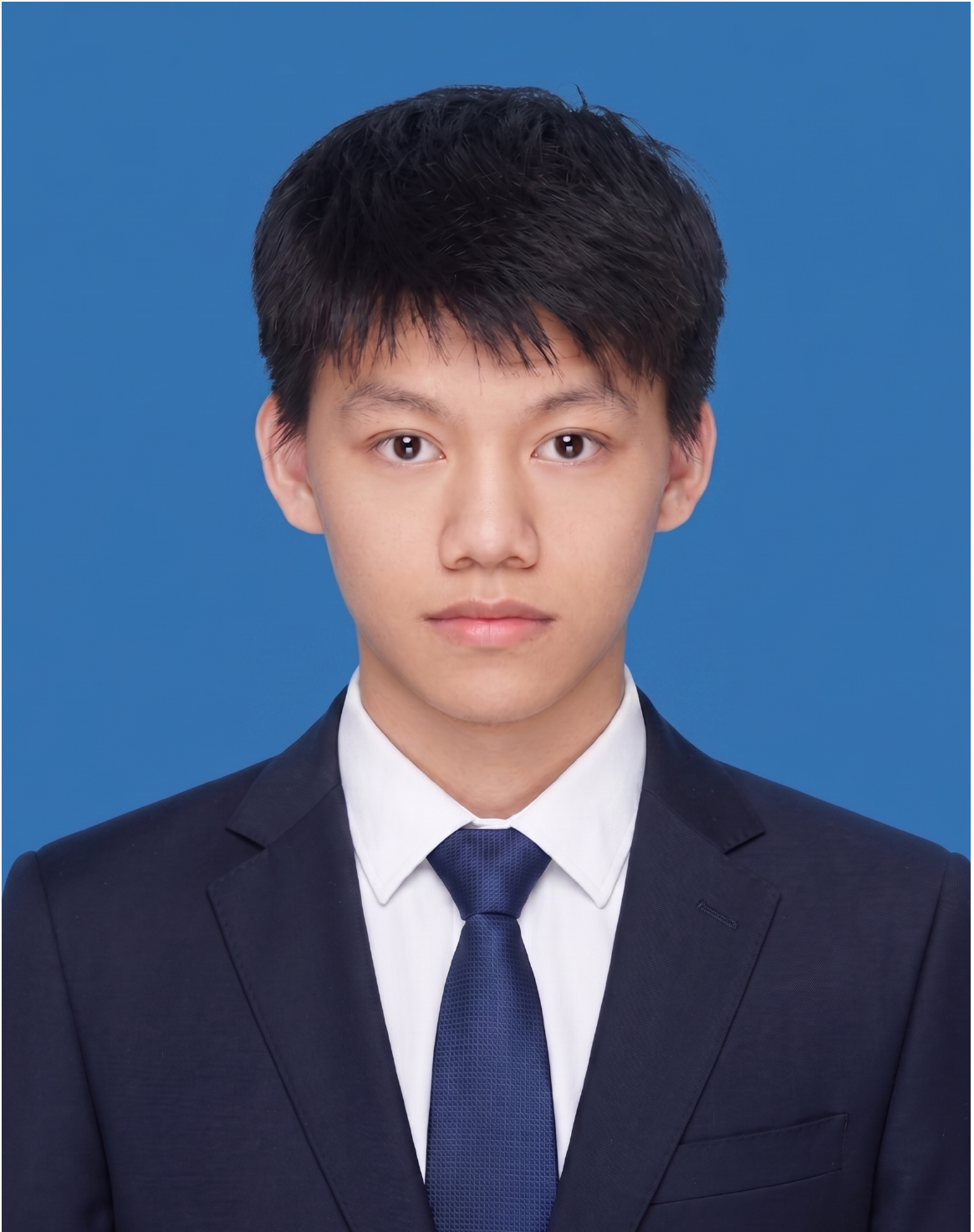}{Aiji Liang}{is currently pursuing a bachelor's degree in the Faculty of Arts and Sciences, Beijing Normal University at Zhuhai, China. His main research interests include multi-agent systems, edge computing, and their applications.}

\authorentry{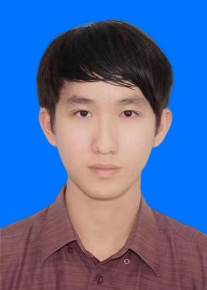}{Jinxi He}{is currently an undergraduate student in the School of Arts and Sciences at Beijing Normal University at Zhuhai, China. His research interests primarily focus on large language models, multi-agent systems, and real-time voice interaction.}

\end{document}